\documentclass[11pt,a4paper]{article}
\usepackage[hyperref]{acl2018}
\usepackage{times}
\usepackage{latexsym}
\usepackage{booktabs}
\usepackage{lscape}
\usepackage{url}
\usepackage{xspace}
\usepackage[colorinlistoftodos,prependcaption,textsize=small]{todonotes}

\usepackage{array,multirow,graphicx}
\newcommand \egrid{\textsc{EGrid}\xspace}
\newcommand \egraph{\textsc{EGraph}\xspace}
\newcommand \lexgraph{\textsc{LexGraph}\xspace}
\newcommand \parseq{\textsc{ParSeq}\xspace}
\newcommand \sentseq{\textsc{SentSeq}\xspace}
\newcommand \clique{\textsc{Clique}\xspace}
\newcommand \sentavg{\textsc{SentAvg}\xspace}
\newcommand \egridconv{\textsc{EGridConv}\xspace}
\newcommand \dataset{\textsc{GCDC}\xspace}

\aclfinalcopy 
\def\aclpaperid{1077} 

\title{Discourse Coherence in the Wild: A Dataset, Evaluation and Methods}

\author{Alice Lai\\
 University of Illinois at Urbana-Champaign\thanks{\ \ Research performed while at Grammarly.}\\
 {\tt aylai2@illinois.edu} \\\And
  Joel Tetreault \\
  Grammarly  \\
  {\tt joel.tetreault@grammarly.com} \\}

\date{}

\begin{document}
\maketitle

\begin{abstract}
To date there has been very little work on assessing discourse coherence methods on real-world data.  To address this, we present a new corpus of real-world texts (\dataset) as well as the first large-scale evaluation of leading discourse coherence algorithms. We show that neural models, including two that we introduce here (\sentavg and \parseq), tend to perform best.  
We analyze these performance differences and discuss patterns we observed in low coherence texts in four domains.

\end{abstract}

\section{Introduction}

Discourse coherence is an important aspect of text quality. It encompasses how  sentences are connected as well as how the entire document is organized to convey information to the reader.
Developing discourse coherence models to distinguish coherent writing from incoherent writing is useful to a range of applications.  An automated coherence scoring model could provide writing feedback, e.g. identifying a missing transition between topics or highlighting a poorly organized paragraph.  Such a model could also improve the quality of natural language generation systems.

One approach to modeling coherence is to model the distribution of entities over sentences. The entity grid \cite{barzilay-lapata:2005:ACL}, based on Centering Theory \cite{Grosz:1995}, was the first of these models. Extensions to the entity grid include additional features \cite{elsner-charniak:2008:ACLShort,elsner-charniak:2011:ACL,feng-lin-hirst:2014:Coling}, a graph representation \cite{guinaudeau-strube:2013:ACL2013,mesgar-strube:2015:*SEM2015}, and neural convolutions \cite{nguyen-joty:2017:ACL}. 
Other approaches have used lexical cohesion \cite{Morris:1991,somasundaran-burstein-chodorow:2014:Coling}, discourse relations \cite{lin-ng-kan:2011:ACL-HLT2011,feng-lin-hirst:2014:Coling}, and syntactic features \cite{louis-nenkova:2012:EMNLP-CoNLL}. Neural networks have also been successfully applied to coherence \cite{li-hovy:2014:EMNLP,nguyen-joty:2017:ACL,li-jurafsky:2017:EMNLP}.
However, until now, these approaches have not been benchmarked on a common dataset.

Past work has focused on the discourse coherence of well-formed texts in domains like newswire \cite{barzilay-lapata:2005:ACL,elsner-charniak:2008:ACLShort} via tasks like sentence ordering that use artificially constructed data. It was unknown how well the best methods would fare on {\em real-world data} that most people generate.



In this work, we seek to address the above deficiencies via four main contributions.  First, we present a new corpus, the Grammarly Corpus of Discourse Coherence (\dataset), for real-world discourse coherence. The corpus contains texts the average person might write, e.g. emails and online reviews, each with a coherence rating from expert annotators (see examples in Table~\ref{tab:examples} and supplementary material). Second, we introduce two simple yet effective neural network models to score coherence.  Third, we perform the first large-scale benchmarking of 7 leading coherence algorithms.  We show that prior models, which performed at a very high level on well-formed and artificially generated data, have markedly lower performance in these new domains.  
Finally, the data, annotation guidelines, and code have all been made public.\footnote{\url{https://github.com/aylai/GCDC-corpus}}




\begin{table*}[htb]
\begin{center}
\begin{footnotesize}
\begin{tabular}{@{}l@{\hspace{1em}}p{14.8cm}@{}}
\toprule
Score & Text \\
\midrule
Low & Should I be flattered? Even a little bit? And, as for my alibi, well, let's just say it depends on the snow and the secret service. So, subject to cross for sure. Do you think there could be copycats? Do you think the guy chose that mask or just picked up the nearest one? Please keep me informed as the case unfolds--
\newline
On another matter, can you believe Dan Burton will be the chair of one of the House subcommittees we'll have to deal w? Irony and satire are the only sane responses.
\newline\
Happy New Year--and here's hoping for many more stories that make us laugh!\\
High & Cheryl,
\newline
I just spoke with Vidal Jorgensen. They expect to be on the ground in about 8 months. They have not yet raised enough money to get the project started -- the total needed is \$6M and they need \$2M to get started. Vidal said they process has been delayed because their work in Colombia and China is consuming all their resources at the moment. Once on the ground, they will target the poorest of the poor and go to the toughest areas of Haiti. They anticipate an average loan size of \$200 and they expect to reach about 10,000 borrowers in five years. They expect to be profitable in 4-5 years.
\newline
Meghann\\
\bottomrule
\end{tabular}
\caption{Examples of texts and coherence scores from the Clinton domain.}
\label{tab:examples}
\end{footnotesize}
\end{center}
\end{table*}

\section{A Corpus for Discourse Coherence}
\label{sec:data}

\subsection{Related Work}

Most previous work in discourse coherence has been evaluated on a sentence ordering task that assumes each text is well-formed and perfectly coherent, and any reordering of the same sentences is less coherent. Presented with a pair of texts -- the original and a random permutation of the same sentences -- a coherence model should be able to identify the original text. More challenging versions of this task (sentence insertion \cite{elsner-charniak:2011:ACL} and paragraph reconstruction \cite{lapata:2003:ACL,li-jurafsky:2017:EMNLP}) all assume that the original text is perfectly coherent.

Datasets for the sentence ordering task tend to use texts that have been professionally written and extensively edited. These have included the Accidents and Earthquakes datasets \cite{barzilay-lapata:2005:ACL}, the Wall Street Journal \cite{elsner-charniak:2008:ACLShort,elsner-charniak:2011:ACL,lin-ng-kan:2011:ACL-HLT2011,feng-lin-hirst:2014:Coling,nguyen-joty:2017:ACL}, and Wikipedia \cite{li-jurafsky:2017:EMNLP}. 



Another task, summary evaluation \cite{barzilay-lapata:2005:ACL}, uses human coherence judgments, but include machine-generated texts. Coherence models are only required to identify which of a pair of texts is more coherent (presumably identifying human-written texts). 

The line of work most closely related to our approach is the application of coherence modeling to automated essay scoring. Essays are written by test-takers, not professional writers, so they are not assumed to be coherent. Manual annotation is required to assign the essay an overall quality score \cite{feng-lin-hirst:2014:Coling} or to rate the coherence of the essay \cite{somasundaran-burstein-chodorow:2014:Coling,burstein:2010:naacl,BursteinTC13}.
While this line of work goes beyond sentence ordering to examine the qualities of a low-coherence text, it has only been applied to test-taker essays. 

In contrast to previous datasets, we collect 
writing from non-professional writers in everyday contexts. Rather than using permuted or machine-generated texts as examples of low coherence, we want to investigate the ways in which people try but fail to write coherently. We present a corpus that contains texts from four domains, covering a range of coherence, each annotated with a document-level coherence score. In Sections \ref{sec:domains}-\ref{sec:agreement}, we describe our data collection process and the characteristics of the resulting corpus. 


\subsection{Domains}
\label{sec:domains}


For a robust evaluation, we selected domains that reflect what an average person writes on a regular basis: forum posts, emails, and product reviews.  For online forum posts, we sampled responses from the Yahoo Answers L6 corpus\footnote{\url{https://webscope.sandbox.yahoo.com/catalog.php?datatype=l}} for the \textbf{Yahoo} domain.  For emails, we used the State Department's release of emails from Hillary Clinton's office\footnote{\url{https://foia.state.gov/Search/Results.aspx?collection=Clinton_Email}} and emails from the Enron Corpus\footnote{\url{https://www.cs.cmu.edu/~./enron/}} to make up our \textbf{Clinton} and \textbf{Enron} domains.  Finally, we sampled reviews of businesses from the Yelp Open Dataset\footnote{\url{https://www.yelp.com/dataset}} for our \textbf{Yelp} domain.

\subsection{Text Selection}

We randomly selected texts from each domain given a few filters.  
We want each text to be long enough to exhibit a range of characteristics of local and global coherence, but not so long that the labeling process is tedious for annotators. Therefore, we considered texts between 100 and 300 words in length. We ignored texts containing URLs (as they often quote writing from other sources) and texts with too many line breaks (usually lists).  


\subsection{Annotation}
\label{sec:annotation}
We collected coherence judgments both from expert raters with prior linguistic annotation experience, as in \citet{burstein:2010:naacl} and from untrained raters via Amazon Mechanical Turk.  This allows us to assess the efficacy of using untrained raters for this task.   We asked the raters to rate the coherence of each text on a 3-point scale from 1 (low coherence) to 3 (high coherence) given the following instructions, which are based on prior coherence annotation efforts \cite{barzilay_CL_2008,BursteinTC13}:
\vspace{-5pt}
\begin{quote}
A text with high coherence is easy to understand, well-organized, and contains only details that support the main point of the text. A text with low coherence is difficult to understand, not well organized, or contains unnecessary details. Try to ignore the effects of grammar or spelling errors when assigning a coherence rating.	
\end{quote}

\paragraph{Expert Rater Annotation} We solicited judgments from 13 expert raters with previous annotation experience. We provided a high-level description of coherence but no detailed rubric, 
as we wanted them to use their own judgment. We also provided examples of low, medium, and high coherence along with a brief justification for each label.
The raters went through a calibration phase during which we provided feedback about their judgments. In the annotation phase, we collected 3 expert rater judgments for each text.


\paragraph{Mechanical Turk Annotation} We collected 5 MTurk judgments for each text from a group of 62 Mechanical Turk annotators who passed our qualification test. 
We again provided a high-level description of coherence. However, we only provided a few examples for each category so as not to overwhelm the annotators.


We were mindful of how the characteristics of each domain might affect the resulting coherence scores. For example, after rating a batch of generally low coherence forum data, business emails may appear to be more coherent. However, our goal is to discover the characteristics of a low coherence business email or a low coherence forum post, not to compare the two domains. Therefore, we recruited new MTurk raters for each domain so as not to bias their scores. The same 13 expert raters worked on all four domains, but we specifically instructed them to consider whether each text was a coherent document \textit{for its domain}.



\subsection{Grammarly Corpus of Discourse Coherence}


The resulting four domains each contain 1200 texts (1000 for training, 200 for testing). Each text has been scored as \{low, medium, high\} coherence by 5 MTurk raters and 3 expert raters. There is one consensus label for the expert ratings and another consensus label for the MTurk ratings. We computed the consensus label by averaging the integer values of the coherence ratings (low = 1, medium = 2, high = 3) over the MTurk or expert ratings and thresholding the mean coherence score (low $\leq 1.8 <$ medium $\leq 2.2 <$ high) to produce a 3-way classification label (Table~\ref{tab:class_distribution}). We observed that the MTurk raters tended to label more texts as ``medium'' coherence than the expert raters. Since the MTurk raters did not go through an extensive training session, they may be less confident in their ratings, defaulting to \textit{medium} as the safe option.

\begin{figure*}[tb]
    \centering
    \includegraphics[width=\textwidth]{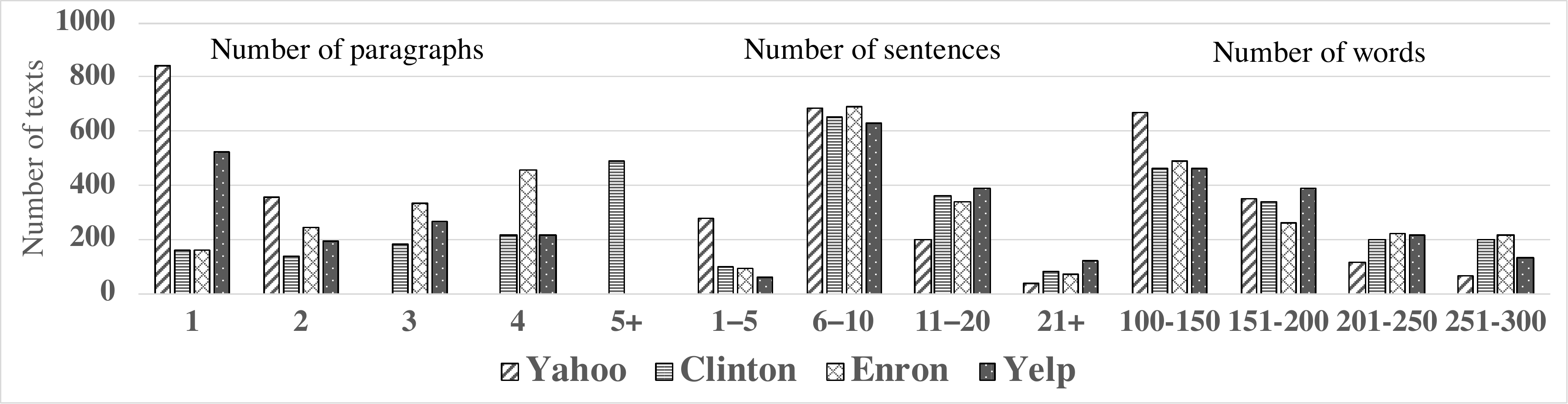}
    \caption{Number of paragraphs, sentences, and words per document.}
    \label{fig:paragraphs}
\end{figure*}

Table~\ref{tab:type_token} contains type and token counts for the full dataset, and Figure~\ref{fig:paragraphs} shows the number of paragraphs, sentences, and words per document.

\begin{table}[htb]
\small
\begin{center}
\begin{tabular}{llrrr}
\toprule
 & & \multicolumn{3}{c}{Coherence Class (\%)} \\
Domain & Raters & Low & Med & High \\
\midrule
Yahoo 	& untrained	& 35.5 & 39.2 & 25.3  \\
	 	& expert	& 46.6 & 17.4 & 37.0 \\
Clinton & untrained & 36.7 & 38.6 & 24.7 \\
		& expert & 28.2 & 20.6 & 51.1 \\
Enron 	& untrained & 34.9 & 44.2 & 20.9 \\
		& expert & 29.9 & 19.4 & 50.7 \\
Yelp 	& untrained & 19.9 & 43.4 & 36.7 \\
		& expert & 27.1 & 21.8 & 51.1 \\
\bottomrule
\end{tabular}
\caption{Distribution of coherence classes as a percentage of the training data.}
\label{tab:class_distribution}
\end{center}
\end{table} 

\begin{table}[htb]
\small
\begin{center}
\begin{tabular}{lrrrr}
\toprule
 & Yahoo & Clinton & Enron & Yelp \\
\# types & 13,235 & 15,564 & 13,694 & 12,201 \\
\# tokens & 189,444 & 220,115 & 223,347 & 213,852 \\
\bottomrule
\end{tabular}
\caption{Type and token counts in each domain.}
\label{tab:type_token}
\end{center}
\end{table} 

\subsection{Annotation Agreement}
\label{sec:agreement}

\begin{table}[htb]
\small
\begin{center}
\begin{tabular}{llrrrr}
\toprule
Domain & Raters & ICC & Weighted $\kappa$ \\
\midrule
Yahoo & untrained & 0.113 $\pm$ 0.024 & 0.060 $\pm$ 0.013 \\
	  & expert & 0.557 $\pm$ 0.010 & 0.386 $\pm$ 0.009 \\
Clinton & untrained & 0.270 $\pm$ 0.020 & 0.156 $\pm$ 0.013 \\
		& expert & 0.398 $\pm$ 0.015 & 0.250 $\pm$ 0.011 \\
Enron	& untrained & 0.141 $\pm$ 0.021 & 0.077 $\pm$ 0.012 \\
		& expert & 0.428 $\pm$ 0.014 & 0.273 $\pm$ 0.011 \\
Yelp 	& untrained & 0.120 $\pm$ 0.026 & 0.069 $\pm$ 0.014 \\
		& expert & 0.304 $\pm$ 0.015 & 0.181 $\pm$ 0.010 \\
\bottomrule
\end{tabular}
\caption{Interannotator agreement (mean and standard deviation) on all domains.}
\label{tab:agreement}
\end{center}
\end{table}

To quantify agreement among annotators, we follow \citet{pavlick_TACL16}'s approach to simulate two annotators from crowdsourced labels. We repeat the simulation 1000 times and report the mean agreement values in Table~\ref{tab:agreement} for both intraclass correlation (ICC) and quadratic weighted Cohen's $\kappa$ for an ordinal scale.

The expert raters have fair agreement \cite{landis} for three of the domains, but agreement among MTurk raters is quite low.
These agreement numbers are the result of an extensive annotation development process and emphasize the difficulty of the task. We recommend that future work in this area leverages raters with a strong annotation background and the time for in-depth instructions. For evaluation, we use the consensus label from the expert judgments. For comparison, we include an experiment using MTurk consensus labels in the supplementary material.  

%
%

\section{Models}
\label{sec:models}


 

We evaluate a range of existing discourse coherence models on \dataset: entity-based models, a word embedding graph model, and neural network models. These models from previous work have been very effective on the sentence ordering task, but have not been used to produce coherence scores. We also introduce two new neural sequence models. 

\subsection{Baseline}

We compute the Flesch-Kincaid grade level \cite{Flesch} of each text and treat it as a coherence score. While Flesch-Kincaid is a readability measure, previous work has treated readability and text coherence as overlapping tasks \cite{barzilay_CL_2008,mesgar-strube:2015:*SEM2015}. For coherence classification, we search over the grade level scores on the training data and select thresholds that result in the highest accuracy. 


\subsection{Entity-based Models}

Entity-based models track entity mentions throughout the text.  
In the majority of our experiments, we applied \citet{barzilay_CL_2008}'s coreference heuristic and consider two nouns to be coreferent only if they are identical. As \citet{elsner-charniak:2011:ACL} noted, automatic coreference resolution often fails to improve coherence modeling results. However, we also evaluate the effect of adding an automatic coreference system in Section~\ref{sec:classification}.

\paragraph{Entity grid (\egrid)}

The entity grid \cite{barzilay-lapata:2005:ACL} is a matrix that tracks entity mentions over sentences. We reimplemented the model from \citet{barzilay_CL_2008}, converting the entity grid into a feature vector that expresses the probabilities of local entity transitions. We use scikit-learn \cite{scikit-learn} to train a random forest classifier over the feature vectors.

\paragraph{Entity graph (\egraph)}

The entity graph \cite{guinaudeau-strube:2013:ACL2013} interprets the entity grid as a graph whose nodes are sentences. Two nodes are connected if they share at least one entity. Graph edges can be weighted according to the number of entities shared, the syntactic roles of the entities, or the distance between sentences. The coherence score of a text is the average out-degree of its graph, so for classification we identify the thresholds that maximize accuracy on the training data.

\paragraph{Entity grid with convolutions (\egridconv)}

\citet{nguyen-joty:2017:ACL} applied a convolutional neural network to the entity grid to capture long-range transitions. We use the authors' implementation.\footnote{\url{https://github.com/datienguyen/cnn_coherence}}

\subsection{Lexical Coherence Graph (\lexgraph)}

The lexical coherence graph \cite{mesgar-strube:2016} represents sentences as nodes of a graph, connecting nodes with an edge if the two sentences contain a pair of similar words (i.e. the cosine similarity of their pre-trained word vectors is greater than a threshold). From the graph, we can extract a feature vector that expresses the frequency of all $k$-node subgraphs. We use the authors' implementation\footnote{\url{https://github.com/MMesgar/lcg}} and train a random forest classifier over the feature vectors.

\subsection{Neural Network Models}

We reimplemented a neural network model of coherence, the sentence clique model, to evaluate its effectiveness on \dataset. We also introduce two new neural network models that are more straightforward to implement than the clique model. 

\paragraph{Sentence clique (\clique)}

\citet{li-jurafsky:2017:EMNLP}'s model operates over cliques of adjacent sentences. For the sentence ordering task, a positive clique is a sequence of $k$ sentences from the original document. A negative clique is created by replacing the middle sentence of a positive clique with a random sentence from elsewhere in the text. The model contains a single LSTM \cite{Hochreiter1997} that takes a sequence of GloVe word embeddings and produces a sentence vector at the final output step. All $k$ sentence vectors are concatenated and passed through a final layer to produce a probability that the clique is coherent. The final coherence score is the average of the scores of all cliques in the document.

We extend \clique to 3-class classification by labeling each clique with the document class label (low, medium, high). To predict the text label, the model averages the predicted coherence class distributions over all cliques.

\paragraph{Sentence averaging (\sentavg)}

To investigate the extent to which sentence order is important in our data, we introduce a neural network model that ignores sentence order. The model contains a single LSTM that produces a sentence vector (the final output vector) from a sequence of GloVe embeddings for the words in that sentence. The document vector is the average over all sentence vectors in that document, and is passed through a hidden layer and a softmax to produce a distribution over coherence labels. 

\paragraph{Paragraph sequence (\parseq)}

The role of paragraph breaks has not been explicitly discussed in previous work. Models like \egrid assume that entity transitions have the same weight whether adjacent sentences $A$ and $B$ occur in the same paragraph or different paragraphs. We expect paragraph breaks to be important for assessing coherence in longer documents.



Therefore, we introduce a paragraph sequence model, \parseq, that can distinguish between paragraphs. \parseq contains three stacked LSTMs: the first takes a sequence of GloVe embeddings to produce a sentence vector, the second takes a sequence of sentence vectors to produce a paragraph vector, and the third takes a sequence of paragraph vectors to produce a document vector. The document vector is passed through a hidden layer and a softmax to produce a distribution over coherence labels. 
A diagram of this model is available in the supplementary material.

\section{Evaluation}
\label{sec:eval}

We evaluate the models on multiple coherence prediction tasks. The best model parameters, reported in the supplementary material, are the result of 10-fold cross-validation over the training data.

For all neural models (\egridconv, \egridconv+coref, \clique, \sentavg, and \parseq), the reported results are the mean of 10 runs with different random seeds, as suggested by \citet{reimers-gurevych:2017}.

We indicate ($\dag$) when the best neural model result is significantly better ($p < 0.05$) than the best non-neural result. We use the one-sample Wilcoxon signed rank test and adjusted the $p$-values to account for the false discovery rate. 

\subsection{Classification}
\label{sec:classification}

\begin{table}
\begin{center}
 \begin{small}
\begin{tabular}{@{}lllll@{}}
	\toprule
    & \multicolumn{4}{c}{Accuracy} \\
	System & Yahoo & Clinton & Enron & Yelp \\	
	\midrule
    Majority class & 41.0 & 55.5 & 44.0 & 54.0 \\
    Baseline & 43.5 & 56.0 & 52.5 & \textbf{55.0} \\ 
    \midrule
	\egrid & 38.0 & 43.0 & 46.0 & 45.5 \\
    \egrid +coref & 41.5 & 48.0 & 47.0 & 49.0 \\
    \egraph & 40.0 & 56.0 & 43.5 & 53.0 \\
    \egraph +coref & 42.5 & 55.0 & 44.0 & 54.0 \\
	\egridconv & 47.0 & 56.3 & 44.8 & 54.2 \\
    \egridconv +coref & 51.0 & 56.6 & 44.7 & 54.0 \\
    \midrule
    \lexgraph & 37.0 & 51.0 & 45.0 & 48.0 \\ 
    \midrule
    \clique & 53.5 & \textbf{61.0}$^{\dag}$ & \textbf{54.4}$^{\dag}$ & 49.1 \\
    \sentavg & 52.6 & 58.4 & 53.2 & 54.3 \\
    \parseq & \textbf{54.9}$^{\dag}$ & 60.2 & 53.2 & 54.4$^{\dag}$ \\	
	\bottomrule
\end{tabular}
 \end{small}
\caption{Three-way classification results on test. 
}
\label{tab:3class_test_appen}
\end{center}
\end{table}

For this task, each text has a consensus label expressing how coherent it is: \{low, medium, high\}. 
We report overall accuracy for all systems on predicting the expert rater consensus label (Table \ref{tab:3class_test_appen}). We repeated this evaluation using the MTurk rater labels and included those results in the supplementary material. 

The neural models outperformed the entity-based and lexical graph models.  Non-neural models showed mixed results, performing on par with or worse than our baseline. Most models perform poorly on Yelp, worse than the baseline, perhaps because Yelp has the lowest annotator agreement among expert raters.

We also tried adding coreference information for the entity-based methods, as it has been shown to be useful in some prior work \cite{barzilay_CL_2008,elsner-charniak:2008:ACLShort}. For the base entity model experiments, we used \citet{barzilay_CL_2008}'s heuristic to determine whether two nouns are coreferent.  For the $+$coref setting, we used the Stanford coreference annotator \cite{clark2015entity} as a preprocessing step before computing the entity grid.  The coreference system yielded consistent performance improvements of 1--5\% accuracy over the corresponding heuristic results, indicating that automatic coreference resolution can help entity-based models in these domains.





\begin{table}
\begin{center}
 \begin{small}
\begin{tabular}{lllll}
	\toprule
    & \multicolumn{4}{c}{Spearman $\rho$} \\
	System & Yahoo & Clinton & Enron  & Yelp \\	
    \midrule
    Baseline & 0.089 & 0.323 & 0.244 & 0.200 \\ 
    \midrule
    \egrid & 0.110 & 0.146 & 0.168 & 0.121 \\
    \egraph & 0.198 & 0.366 & 0.074 & 0.103 \\
	\egridconv & 0.204 & 0.251 & 0.258 & 0.104 \\ 
    \midrule
    \lexgraph & 0.130 & 0.049 & 0.273 & $-0.008$ \\ 
    \midrule
    \clique & 0.474 & 0.474 & 0.416 & 0.304 \\
    \sentavg & 0.466 & \textbf{0.505}$^{\dag}$ & 0.438 & 0.311 \\	
    \parseq & \textbf{0.519}$^{\dag}$ & 0.448 & \textbf{0.454}$^{\dag}$ & \textbf{0.329}$^{\dag}$ \\	
	\bottomrule
\end{tabular}
 \end{small}
\caption{Score prediction results on test.}
\label{tab:score_predict_test}
\end{center}
\end{table}

\subsection{Score Prediction}


A 3-point coherence score might not reflect the range of coherence that actually exists in the data. We can instead present a more fine-grained score prediction task where the gold score is the mean of the three expert rater judgments (low coherence = 1, medium = 2, high = 3). 
In Table~\ref{tab:score_predict_test}, we report Spearman's rank correlation coefficient between the gold scores and the predicted coherence scores.  As in the classification task, the neural methods convincingly outperformed all other methods, with \parseq the top performer in three out of four domains.


\subsection{Sentence Ordering}
The sentence ordering ranking task is a somewhat artificial evaluation, as a document whose sentences have been randomly shuffled does not resemble a human-written text that is not very coherent. However, we still want to assess whether good performance on previous sentence ordering datasets translates to \dataset. Since the sentence ordering task assumes well-formed texts, we use only the high coherence texts. As a result, there are fewer texts than for the classification task, as we show below. The number of training examples is 20 times the number of texts, as we generate 20 random permutations for each text.

\begin{center}
\begin{small}
\begin{tabular}{lrrrr}
\toprule
 & Yahoo & Clinton & Enron & Yelp \\
\midrule
Train texts & 369 & 511 & 507 & 511\\
Test texts & 76 & 111 & 88 & 108 \\
\bottomrule
\end{tabular}
\end{small}
\end{center}





Table \ref{tab:permutation_test} shows the accuracy of each system on identifying the original text in each (original, permuted) text pair. We leave out the baseline and \sentavg because they ignore sentence order.  We also simplify \parseq to a sentence sequence model (\sentseq) containing only two LSTMs because the sentence ordering task ignores paragraph information.  As in the prior two evaluations, the neural models perform best in most domains, although \egraph is best on Yahoo.

\begin{table}
\begin{center}
\begin{small}
\begin{tabular}{@{}lllll@{}}
	\toprule
    & \multicolumn{4}{c}{Accuracy} \\
	System & Yahoo & Clinton & Enron & Yelp \\	
	\midrule
    Random baseline & 50.0 & 50.0 & 50.0 & 50.0 \\
    \midrule
	\egrid & 55.9 & 78.2 & 77.4 & 62.9 \\
    \egraph & \textbf{64.0} & 75.3 & 75.9 & 59.5 \\
    \egridconv & 54.8 & 75.5 & 73.1 & 58.7 \\
	\midrule
    \lexgraph & 62.5 & 78.3 & 77.9 & 60.8 \\ 
    \midrule
        \clique & 57.8 & \textbf{89.4}$^{\dag}$ & \textbf{88.7}$^{\dag}$ & 64.6 \\
    \sentseq & 58.3 & 88.0 & 87.1 & \textbf{74.2}$^{\dag}$ \\

	\bottomrule
\end{tabular}
\end{small}
\caption{Sentence ordering results on test data.}
\label{tab:permutation_test}
\end{center}
\end{table}

\begin{table}[t]
\begin{center}
\begin{footnotesize}
\begin{tabular}{@{}lllll@{}}
	\toprule
	System & Yahoo & Clinton & Enron  & Yelp \\	
    \midrule
    Baseline & 0.283 & 0.255 & 0.341 & 0.197 \\ 
    \midrule
    \egrid & 0.258 & 0.260 & 0.294 & 0.161 \\
    \egraph & 0.308 & \textbf{0.382} & 0.278 & 0.117 \\
	\egridconv & 0.360 & 0.238 & 0.279 & 0.169 \\ 
    \midrule
	\lexgraph & 0.342 & 0.094 & 0.357 & 0.000 \\
    \midrule
    \clique & 0.055 & 0.000 & 0.077 & 0.146 \\
    \sentavg & \textbf{0.481}$^{\dag}$ & 0.332  & \textbf{0.393}$^{\dag}$ & \textbf{0.199} \\	
    \parseq & 0.447 & 0.296 & 0.373 & 0.112 \\	
	\bottomrule
\end{tabular}
\end{footnotesize}
\caption{Minority class predictions, $F_{0.5}$ score on test data.}
\label{tab:minority_test_f}
\end{center}
\end{table}

\subsection{Minority Class Classification}

One application of a coherence classification system would be to provide feedback to writers by flagging text that is not very coherent. 
Such a system should identify only the most incoherent areas of the text, to ensure that the feedback is not a false positive. To evaluate this scenario, we present a minority class classification problem where only 15-20\% of the data is low coherence:

\begin{center}
\begin{small}
\begin{tabular}{lrrrr}
	\toprule
	 & Yahoo & Clinton & Enron & Yelp \\	
    Low coherence \% & 30.0 & 16.6 & 18.4 & 14.8 \\
	\bottomrule
\end{tabular}
\label{tab:minority_class_percent}
\end{small}
\end{center}

We relabel a text as \textit{low coherence} if at least two expert annotators judged the text to be low coherence, and relabel as \textit{not low coherence} otherwise.


We report the $F_{0.5}$ score of the low coherence class in Table~\ref{tab:minority_test_f}, where precision is emphasized twice as much as recall.\footnote{Precision and recall are in the supplementary material.}   This is in line with evaluation standards in other writing feedback applications \cite{ng-EtAl:2014:W14-17}.
Again, the neural models perform best in most domains. However, the results of this experiment in particular show that there is still a large gap between the performance of these models and what might be required for high-precision real-world applications.

\subsection{Cross-Domain Classification}

Up to this point, we assumed that the four domains are different enough from one another that we should train separate models for each. To test this assumption, we train \parseq, one of the top performing neural models, in one domain (e.g. Yahoo) and evaluate it in a different domain (Clinton, Enron, and Yelp).
Table~\ref{tab:cross_domain_parseq_test} compares the in-domain results (the diagonal) to the cross-domain results. 

\begin{table}
\begin{center}
\begin{small}
\begin{tabular}{clrrrr}
\toprule
& & \multicolumn{4}{c}{\textbf{Test}} \\
& & Yahoo & Clinton & Enron & Yelp \\
\multirow{3}{*}{\rotatebox[origin=c]{90}{\parbox[c]{1.5cm}{\centering \textbf{Train}}}} & Yahoo & \textbf{54.9} & 56.7 & 50.6 & \textbf{55.3} \\
				& Clinton & 51.8 & \textbf{60.2} & 50.7 & 40.4 \\
				& Enron & 51.5 & 59.9 & \textbf{53.2} & 50.8 \\
				& Yelp & 48.3 & 55.5 & 44.0 & 54.4 \\
\bottomrule
\end{tabular}
\caption{Cross-domain accuracy of \parseq on three-way classification test data.}
\label{tab:cross_domain_parseq_test}
\end{small}
\end{center}
\end{table}

While the model's accuracy generally decreases when transferred to a different domain, sometimes this decrease is not too severe: for example, training on Yahoo/Enron data and testing on Clinton data, or training on Yahoo data and testing on Yelp data. It is reasonable that training on one set of business emails (Clinton or Enron) produces a model that can accurately score the coherence of other sets of business emails. Similarly, both Yahoo and Yelp contain online text written for public consumption which may share coherence characteristics, so it is not surprising that a model trained on Yahoo data works on Yelp (even outperforming the Yelp-trained model).

These results indicate that we might be able to train a better coherence model by combining all our data across multiple domains. We evaluate this theory in Table~\ref{tab:concat_domain}, comparing the results of the \parseq model evaluated in-domain (e.g. trained and tested on Yahoo data) to a model trained on the combined training data from all four domains. With four times as much training data, the performance of \parseq improves in all domains, indicating that better coherence models may be trained from data outside of a specific, narrow domain.

\begin{table}
\begin{center}
\begin{small}
\begin{tabular}{lrrrrr}
\toprule
& \multicolumn{4}{c}{\textbf{Test accuracy}} \\
& Yahoo & Clinton & Enron & Yelp \\
\midrule
Train in-domain & 54.9 & 60.2 & 53.2 & 54.4 \\
Train all data & \textbf{58.5} & \textbf{61.0} & \textbf{53.9} & \textbf{56.5}  \\
\bottomrule
\end{tabular}
\caption{Classification accuracy of \parseq when trained on data from all four domains. }
\label{tab:concat_domain}
\end{small}
\end{center}
\end{table}

\subsection{Discussion}  
\label{sec:results}

We observe some trends across our experiments. The basic entity models (\egrid and \egraph) tend to perform poorly, often barely outperforming the baseline. The entity grids computed from \dataset texts are often extremely sparse, so meaningful entity transitions between sentences are infrequent. In addition, scoring the coherence of a text (either classification or score prediction) is more difficult than the sentence ordering task, where basic entity models do outperform the random baseline by a reasonable margin. Both the data and the difficulty of the tasks contribute to poor performance from the basic entity models.

%

The neural network models almost always outperform other models. This supports \citet{li-jurafsky:2017:EMNLP}'s claim that neural models are better able to extend to other domains compared to previous coherence models. Our \parseq and \sentavg models are easier to implement than \clique and outperform \clique on a majority of experiments. \egridconv usually does not perform as well as the other neural models, but it usually improves over \egrid.

Finally, the relative success of \sentavg, which ignores sentence order, is evidence that identifying a document's original sentence order is not the same as distinguishing low and high coherence documents. The large number of parameters in \parseq may explain why it is sometimes outperformed by \sentavg.


%
%

\section{Analysis}
\label{sec:analysis}

To better understand what distinguishes a low coherence text from a high coherence text, we manually analyzed Yahoo and Clinton texts whose labels were unanimously agreed on by all three raters.
Regardless of the domain, many low coherence texts are not well-organized and appear to be written almost as stream of consciousness. They often lack connectives, resembling a list of points rather than a coherent document.

Incoherent Yahoo texts often contain extremely long sentences, lack paragraph breaks, and veer off-topic without a transition or any connection back to the main point. This is an especially frequent occurrence with personal anecdotes.

Low coherence Clinton emails make better use of paragraphs, but they too often lack transitions between topics. In addition, missing information was a primary reason for low coherence scores. We provided the raters with individual emails, not the entire email thread, so raters had less information than the original recipient of the email. This amplifies the detrimental effects on coherence of jargon, abbreviation, and missing context. However, overuse of these compression strategies can result in low coherence even for the intended recipient, so it is worth modeling their effects.

Across domains, coherent texts have a clear topic that is maintained throughout the text, and they are well-organized, with sentences, paragraphs and sub-topics following a logical ordering. 
Connectives, such as \textit{however}, \textit{for example}, \textit{in turn}, \textit{also}, \textit{in addition} are used more frequently to assist the structure and flow.  

Although sentence order is clearly important, rewriting a disorganized text is not as simple as reordering sentences. Even if changing the location of one sentence increases coherence, a true fix would still require rewriting that sentence or the surrounding sentences. Our analysis indicates that the sentence reordering task is not a good evaluation of whether models can truly be useful to the task of identifying low coherence texts.

\section{Conclusion}
\label{sec:conclusion}
In this paper, we examine the evaluation of discourse coherence by presenting a new corpus (\dataset) to benchmark leading methods on real-world data in four domains.  While neural models outperform others across multiple evaluations, much work remains before any of these methods can be used for real-world applications. 
That said, our \sentavg and \parseq models serve as simple and effective methods to use in future work.

We recommend that future evaluations move away from the sentence ordering task.  While it is an easy evaluation to carry out, the performance numbers overpredict the success of those systems in real-world conditions.  For example, prior evaluations \cite{nguyen-joty:2017:ACL,li-jurafsky:2017:EMNLP} report performance numbers around or above 90\% accuracy, which contrasts with the much lower figures shown in this paper.
In addition, we recommend that future annotation efforts leverage expert raters, preferably with a background in annotation, as this task is difficult for untrained workers on crowdsourcing platforms.

By releasing \dataset, the annotation guidelines, and our code, we hope to encourage future work on more realistic coherence tasks.

\section*{Acknowledgments}

The authors would like to thank Yahoo Research and Yelp for making their data available, and Jiwei Li and Mohsen Mesgar for sharing their code. Thanks also to Michael Strube, Annie Louis, Rebecca Hwa, Dimitrios Alikaniotis, Claudia Leacock, Courtney Napoles, Jill Burstein, Mirella Lapata, Martin Chodorow, Micha Elsner, and the anonymous reviewers for their helpful comments. 

\bibliography{acl2018}
\bibliographystyle{acl_natbib}

\newpage
\appendix
\section{Supplementary Material}
\label{sec:supplementary}

\subsection{Corpus Examples}

Table~\ref{tab:examples} contains additional examples of texts from our corpus, specifically from the Yahoo Answers domain, with their coherence labels.
 
\begin{table*}[htb]
\begin{center}
\begin{footnotesize}
\begin{tabular}{@{}l@{\hspace{1em}}l@{\hspace{1em}}p{13.5cm}@{}}
\toprule
Domain & Score & Text \\
\midrule
Yahoo & Low & I see it, but then again almost every war entered by the U.S. is connected to gaining something. The U.S. is just using politically correct was of taking over a country without anybody noticing it. They enter a war and some how we come out better than the country we went in to help. We say we are helping but if the country has nothing for us then we don't bother with it. For example: Korea stated and I quote ``we have nuclear weapons and we plan to use them'' so how come we are in Iraq who have no weapons? Well maybe the U.S. sees no threat but then again somebody did sneak into the country and take over planes. Also not to long ago it was common for somebody to hijack a plane. Well that is all I have to say on the matter.\\
Yahoo & High & Don't be intimidated by Impressionism.  It is simply a style worked in loose strokes.  The idea is to give an ``impression'' of the subject.  Choose a simple subject, like a still life or bowl of fruit.  Then layout your palette using the colors you see (make sure to look for subtle colors only an artist might see...such as the ``blue'' in an apple), and with a larger than usual brush, stroke the basic shapes in a medium value, then add shadows, then a highlight layer.  That should do for a class project in Impressionism.  The danger would come from over-working the painting.  You don't want fine strokes or details, remember just the ``impression'' of your subject.  The whole idea is to stay loose and free.  A lot of people struggle with it.  The trick is to just paint without worrying too much.  Good luck.\\
\bottomrule
\end{tabular}
\caption{Examples of texts with coherence scores.}
\label{tab:examples}
\end{footnotesize}
\end{center}
\end{table*}

\subsection{Annotator Instructions}

The annotation instructions in Section 2.4 are the simplified instructions that we provided to Mechanical Turk workers. The expert annotators received a longer version of those instructions, which are available in Table~\ref{tab:expert_instructions}.

\begin{table*}
\centering
\begin{small}
\begin{tabular}{@{}p{16cm}@{}}	
You will be given a short text (100-300 words) to read. We will specify which one of several domains the text comes from, and in some domains we will provide additional context for the text.\\[4pt]

Your task is to rate the coherence of the text from 1 to 3 (1 means low coherence, 3 means high coherence).\\[4pt]

Coherence in writing refers to how well ideas flow from one sentence to the next, and from one paragraph to the next. A text that is highly coherent is easy to understand and easy to read. This usually means the text is well-organized, logically structured, and presents only information that supports the main idea. On the other hand, a text with low coherence is difficult to understand. This may be because the text is not well organized, contains unrelated information that distracts from the main idea, or lacks transitions to connect the ideas in the text.\\[4pt]

Try to ignore the effects of grammar or spelling errors when assigning a coherence rating, as long as the errors do not significantly interfere with your ability to read and understand the text. In the email data, assume that jargon and acronyms are used correctly, and do your best to judge coherence despite that.\\[4pt]

You should assign a coherence rating to the text based on whether it is a coherent example of text \textit{in that domain}. A reader has different expectations about how a business email should be written compared to a post on an online forum, and the coherence rating should reflect this difference. A business email with a score of 1 is not necessarily incoherent in the same way that a very incoherent Yahoo Answers post is, but it is not very coherent \textit{for a business email}. 
\end{tabular}
\end{small}
\caption{The annotation instructions we provided to expert annotators.}
\label{tab:expert_instructions}
\end{table*}

\subsection{Model Details}

\begin{figure}[h]
    \centering
    \includegraphics[width=.45\textwidth]{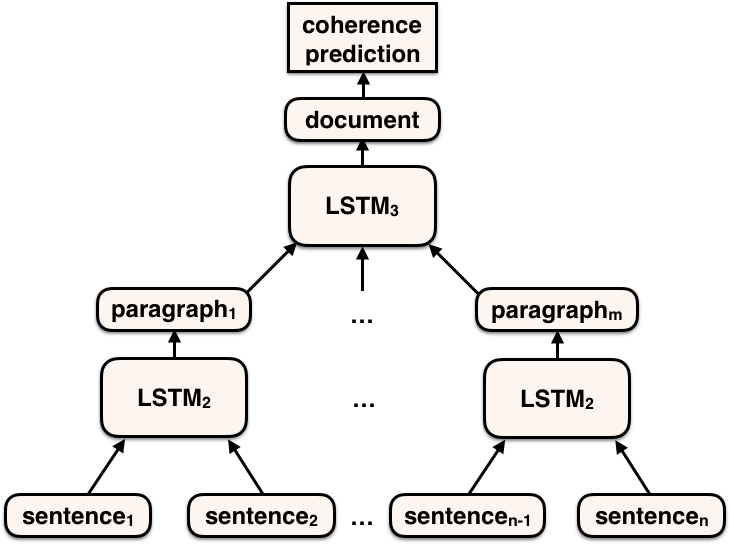}
    \caption{Structure of \parseq model. The sentence vectors are the output from the first LSTM (not pictured), which takes GloVe word embeddings as input.}
    \label{fig:model_diagram}
\end{figure}

Figure~\ref{fig:model_diagram} shows the structure of \parseq. The sentence vectors pictured are the output at the final timestep from the first LSTM (not pictured), which takes GloVe word embeddings as input. A second LSTM takes these sentence vectors as input and produces paragraph vectors, and a third LSTM takes a sequence of paragraph vectors and produces a single document vector.

\begin{table}
\begin{center}
 \begin{small}
\begin{tabular}{lrrrr}
	\toprule
    & \multicolumn{4}{c}{Accuracy} \\
	System & Yahoo & Clinton & Enron & Yelp \\	
	\midrule
    Majority class & 39.5 & 40.5 & 44.0 & 40.5 \\
    Baseline & 35.0 & 43.5 & 45.0 & 41.5 \\ 
    \midrule
	\egrid & 43.0 & 41.0 & 45.5 & 43.0 \\
    \egraph & 39.5 & 41.5 & 44.5 & 40.5 \\
	\egridconv & 41.0 & 43.5 & 44.5 & 54.0 \\
    \lexgraph & 38.0 & 36.0 & 48.0 & 45.5 \\
    \clique & 48.0 & 45.0 & 52.5 & 51.0 \\
    \sentavg & \textbf{52.0} & 48.5 & 55.5 & 49.0 \\	
    \parseq & 47.5 & \textbf{51.0} & \textbf{56.5} & \textbf{57.5} \\
	\bottomrule
\end{tabular}
 \end{small}
\caption{Three-way classification results on test data. Untrained rater judgments.}
\label{tab:3class_test_mturk}
\end{center}
\end{table}

\subsection{Additional Results}

Table~\ref{tab:3class_test_mturk} contains the classification test results of all systems when the consensus labels come from the Mechanical Turk judgments rather than the expert judgments. 

\begin{table}
\begin{center}
\begin{footnotesize}
\begin{tabular}{@{}lrrrrrrrr@{}}
	\toprule
	System & \multicolumn{2}{c}{Yahoo} & \multicolumn{2}{c}{Clinton} & \multicolumn{2}{c}{Enron}  & \multicolumn{2}{c}{Yelp} \\	
    & p & r & p & r & p & r & p & r \\
    \midrule
    Baseline & 25 & 61 & 26 & 24 & 33 & 38 & 17 & 42  \\ 
    \midrule
    \egrid & 31 & 16 & 36 & 12 & 57 & 10 & 33 & 5 \\
    \egraph & 26 & 94 & 35 & 58 & 25 & 45 & 10 & 68  \\
	\egridconv & 31 & 41 & 16 & 24 & 22 & 40 & 50 & 05 \\ 
    \midrule
    \lexgraph & 26 & 29 & 20 & 3 & 55 & 15 & 0 & 0  \\ 
    \midrule
    \clique & 7 & 3 & 0 & 0 & 17 & 3 & 100 & 05 \\
    \sentavg & 38 & 73 & 39 & 36 & 42 & 33 & 36 & 21 \\	
    \parseq & 43 & 51 & 21 & 39 & 57 & 20 & 13 & 11  \\	
	\bottomrule
\end{tabular}
\end{footnotesize}
\caption{Minority class predictions, precision/recall results on test data.}
\label{tab:minority_test}
\end{center}
\end{table}

Table~\ref{tab:minority_test} contains the precision and recall results for the minority class classification test. For neural models, we report precision and recall for one run on test (F0.5 scores in Section 4.4 were averaged over 10 runs).

\begin{table}
\begin{center}
 \begin{small}
\begin{tabular}{lr}
	\toprule
	System & Accuracy \\	
	\midrule
    Random baseline & 50.0 \\
    \midrule
	\egrid & 83.0 \\
    \egraph & 65.7 \\
    \egridconv & 82.2\\
    \lexgraph & 72.7 \\
    \clique & 60.9 \\
    \sentseq & 74.1 \\
	\bottomrule
\end{tabular}
\end{small}
\caption{Sentence ordering results on WSJ test data.}
\label{tab:wsj_perm_test}
\end{center}
\end{table}

To compare all models on an established dataset, we report results on the sentence ordering task using the Wall Street Journal (WSJ) portion of the Penn Treebank. Following previous work, we use 20 random permutations of each article and the train/test split defined by Tien Nguyen and Joty (2017) (train = Section 00-13, test = 14-24). Table~\ref{tab:wsj_perm_test} contains the results of all models on WSJ. These results verify our re-implementation of the \egrid model, as well as establishing the reasonable performance of our neural sequence model on news text.


\subsection{Model Parameters}

We specify the parameters for all models and experiments in Tables \ref{tab:param_vals1} and \ref{tab:param_vals2}. Additionally, for the combined training data experiment (Table 10 in the paper), we train parseq with LSTM dimensionality = 100, hidden layer = 200, dropout = 0.5. 

\paragraph{\egrid} \textit{Sequence length} is the length of the transition sequences used to compute the feature vector from the entity grid. For salience, we follow Barzilay and Lapata (2008) and split entities into two salience classes (doubling the number of features) based on whether their frequency is greater than the \textit{salience threshold}. (Salience = off means that there is only one salience class containing all entities.) \textit{Syntax} indicates whether we consider grammatical roles (subject, object, other) in building the entity grid.

\paragraph{\egraph} The \textit{graph type} specifies whether we use an unweighted graph (u), a graph weighted by the number of entities shared between sentences (w), or a graph weighted by syntactic role information (syn). \textit{Distance} indicates whether edge weights are decreased according to the distance between sentences. 

\paragraph{\egridconv} We specify dropout rate, batch size, and entity role embedding size. For the convolution layer, we specify filter number, window size, and pooling length.

\paragraph{\lexgraph} We define the similarity \textit{threshold} used to filter out edge weights between sentences, and \textit{k} as the size of the subgraphs we consider when extracting features from the document graph.

\paragraph{\clique} We define the dropout rate, the LSTM dimensionality, and the hidden layer dimensionality. \textit{Window size} is the number of sentences in a clique.

\paragraph{\sentavg, \parseq} For both models, we specify the dropout rate, the LSTM dimensionality, and the hidden layer dimensionality. For \parseq, the LSTM dimensionality applies to all 3 LSTMs.

\begin{table*}
\begin{center}
\begin{footnotesize}
\begin{tabular}{llrrrrrrrr}
\toprule
 & & \multicolumn{4}{c}{\textbf{Classification}} & \multicolumn{4}{c}{\textbf{Score Prediction}} \\
Model & Parameter & Yahoo & Clinton & Enron & Yelp & Yahoo & Clinton & Enron & Yelp \\
\midrule
Baseline & threshold1 & 6.5 & 6.5 & 6.0 & 2.5 & -- & -- & -- & --  \\
			& threshold2 & 7.0 & 7.0 & 6.5 & 3.0 & -- & -- & -- & -- \\
\midrule
\egrid	& sequence length & 4 & 3 & 4 & 2 & 2 & 2 & 4 & 3 \\
		& salience threshold & off & 2 & 4 & 4 & 2 & off & 3 & 2 \\
        & syntax & on & off & on & on & off & off & on & on \\
        \midrule
\egraph 	& graph type & syn & syn & syn & syn & u & w & w & syn \\
		& distance & no & no & no & no & yes & yes & yes & no \\
        & threshold1 & 15.0 & 0.1 & 0.1 & 0.5 & -- & -- & -- & -- \\
        & threshold2 & 16.0 & 1.1 & 1.1 & 1.6 & -- & -- & -- & -- \\
        \midrule
\egridconv 	& dropout & 0.2 & 0.2 & 0.5 & 0.2  & 0.2 & 0.5 & 0.5 & 0.2 \\
			& filter & 100 & 100 & 100 & 200 & 200 & 200 & 200 & 100 \\
            & window & 4 & 2 & 2 & 6 & 2 & 2 & 2 & 4  \\
            & pool & 3 & 7 & 3 & 5 & 5 & 3 & 3 & 3 \\
            & batch & 128 & 128 & 32 & 128 & 32 & 32 & 32 & 32 \\
            & embedding size & 100 & 100 & 100 & 200 & 100 & 200 & 200 & 100 \\
            \midrule
\lexgraph	& threshold & 0.7 & 0.5 & 0.7 & 0.9 & 0.5 & 0.3 & 0.7 & 0.9  \\
	& k & 6 & 6 & 6 & 5 & 6 & 6 & 4 & 5 \\
    \midrule
\clique	& dropout & 0.5 & 0.5 & 0.5 & 0.5 & 0.5 & 0.5 & 0.5 & 0.5 \\
		& LSTM dim & 100 & 100 & 200 & 100 & 100 & 100 & 200 & 100 \\
        & hidden dim & 200 & 200 & 200 & 200 & 200 & 200 & 200 & 100  \\
        & window size & 3 & 3 & 3 & 7 & 7 & 7 & 7 & 4 \\
\midrule
\sentavg 	& dropout & 0.5 & 0.5 & 0.5 & 0.2 & 0.2 & 0.5 & 0.5 & 0.2  \\
			& LSTM dim & 200 & 50 & 200 & 50 & 300 & 300 & 300 & 300 \\
           	& hidden dim & 200 & 50 & 100 & 300 & 100 & 100 & 50 & 50 \\
            \midrule
\parseq	& dropout & 0.5 & 0.5 & 0.5 & 0.5 & 0.5 & 0.2 & 0.5 & 0.2 \\
		& LSTM dim & 200 & 300 & 50 & 50 & 300 & 200 & 100 & 300  \\
        & hidden dim & 100 & 100 & 100 & 200 & 100 & 50 & 100 & 100 \\
\bottomrule
\end{tabular}
\end{footnotesize}
\caption{Best parameter values for classification and score prediction experiments.}
\label{tab:param_vals1}
\end{center}
\end{table*} 

\begin{table*}[htb]
\begin{center}
\begin{footnotesize}
\begin{tabular}{llrrrrrrrrr}
\toprule
 & & \multicolumn{5}{c}{\textbf{Sentence Ordering}} & \multicolumn{4}{c}{\textbf{Minority Class}} \\
Model & Parameter & Yahoo & Clinton & Enron & Yelp & WSJ & Yahoo & Clinton & Enron & Yelp \\
\midrule
Baseline & threshold1  & -- & -- & -- & -- & -- & 8.0 & 6.5 & 6.0 & 5.0 \\
			& threshold2 & -- & -- & -- & -- & -- & -- & -- & -- & -- \\
\midrule
\egrid	& sequence length &  4 & 4 & 4 & 4 & 3 & 2 & 2 & 2 & 3 \\
		& salience threshold & 4 & off & 4 & off & 4 & off & off & 2 & 2 \\
        & syntax & on & on & off & on & on & off & off & on & off \\
        \midrule
\egraph 	& graph type &  syn & w & w & w & w & u & w & w & w \\
		& distance & yes & yes & yes & yes & yes & yes & yes & yes & no \\
        & threshold1 & -- & -- & -- & -- & -- & 1.2 & 0.5 & 0.9 & 2.2 \\
        & threshold2 & -- & -- & -- & -- & -- & -- & -- & -- & --\\
        \midrule
\egridconv 	& dropout & 0.2 & 0.2 & 0.2 & 0.2 & 0.5 & 0.2 & 0.5 & 0.5 & 0.5 \\
			& filter & 100 & 100 & 100 & 100 & 150 & 100 & 200 & 200 & 200 \\
            & window & 6 & 6 & 4 & 6 & 6 & 2 & 4 & 6 & 6 \\
            & pool & 7 & 7 & 7 & 7 & 6 & 3 & 3 & 5 & 7 \\
            & batch & 32 & 32 & 32 & 128 & 128 & 128 & 32 & 32 & 32 \\
            & embedding size & 200 & 100 & 200 & 100 & 100 & 100 & 200 & 100 & 200 \\
            \midrule
\lexgraph	& threshold & 0.9 & 0.9 & 0.9 & 0.9 & 0.3 & 0.5 & 0.7 & 0.5 & 0.9 \\
	& k &  4 & 3 & 4 & 4 & 4 & 4 & 3 & 3 & 3 \\
    \midrule
\clique	& dropout & 0.5 & 0.2 & 0.2 & 0.2 & 0.2 & 0.2 & 0.2 & 0.5 & 0.2 \\
		& LSTM dim & 100 & 100 & 100 & 300 & 300 & 50 & 50 & 50 & 50 \\
        & hidden dim & 100 & 100 & 50 & 50 & 50 & 50 & 300 & 50 & 200 \\
        & window size &  7 & 5 & 5 & 5 & 7 & 5 & 7 & 5 & 7 \\
\midrule
\sentavg 	& dropout & -- & -- & -- & -- & -- & 0.2 & 0.5 & 0.2 & 0.2 \\
			& LSTM dim & -- & -- & -- & -- & -- & 200 & 200 & 50 & 200 \\
           	& hidden dim & -- & -- & -- & -- & -- & 300 & 200 & 50 & 50 \\
            \midrule
\parseq	& dropout & 0.5 & 0.2 & 0.2 & 0.2 & 0.5 & 0.5 & 0.2 & 0.5 & 0.2 \\
		& LSTM dim & 50 & 300 & 300 & 300 & 300 & 100 & 50 & 200 & 300 \\
        & hidden dim & 200 & 300 & 200 & 100 & 200 & 50 & 300 & 50 & 100 \\
\bottomrule
\end{tabular}
\end{footnotesize}
\caption{Best parameter values for sentence ordering and minority class classification experiments.}
\label{tab:param_vals2}
\end{center}
\end{table*}

\end{document}